%% file: on2v_sdm.tex
\documentclass[twoside,leqno,twocolumn]{article}
\usepackage{ltexpprt}

\usepackage{amssymb}
\usepackage{graphicx}

\usepackage{url}
\usepackage{color}
\usepackage{comment}

\usepackage{booktabs}
\usepackage{amsfonts}
\usepackage{amsmath}
\usepackage{amssymb}
\usepackage{makecell}
\usepackage{multirow}
\usepackage{pbox}
\usepackage{mathtools}
\usepackage[ruled, vlined, linesnumbered, noend]{algorithm2e}
\usepackage{caption}
\usepackage{tikz}
\usepackage{subfigure}
\usepackage{comment}
\usepackage{enumitem}

\usepackage[flushleft]{threeparttable}
\usepackage{afterpage}

\usepackage{graphicx,xspace}
\usepackage{epstopdf}
\usepackage{stfloats}
\usepackage{setspace}

\def\inv{\vspace{-0.1cm}}

\def\bhline{\specialrule{.2em}{0em}{0em}}
\newcommand{\bigO}[1]{{\rm O} (#1)\xspace}
\newcommand{\stitle}[1]{\vspace{0.3ex}\noindent{\bf #1}}
\newcommand{\ontovec}{{\texttt{On2Vec}}}

\newcommand*\samethanks[1][\value{footnote}]{\footnotemark[#1]}

\begin{document}
\title{\ontovec: Embedding-based Relation Prediction for\\ Ontology Population}
\author{Muhao Chen\thanks{Dept. of Comp. Sci., University of California, Los Angeles. \{muhaochen, shirleychen, xuezijun, zaniolo\}@cs.ucla.edu}\\
\and
Yingtao Tian\thanks{Stony Brook University. yittian@cs.stonybrook.edu} \\
\and
Xuelu Chen\samethanks[1] \\
\and
Zijun Xue\samethanks[1] \\
\and
Carlo Zaniolo\samethanks[1]}
\date{}
\maketitle

\inv
\input{abstract}
\inv
\input{introduction}
\input{related_work}
\input{model}

\input{experiment}
\input{conclusion}

\fancyfoot[R]{\footnotesize{\textbf{Copyright \textcopyright\ 2018 by SIAM\\
Unauthorized reproduction of this article is prohibited}}}

\bibliographystyle{siamplain}
\begingroup
\bibliography{ref}
\endgroup
\end{document}

%% file: abstract.tex
\inv
\begin{abstract}
\inv
Populating ontology graphs represents a long-standing problem for
the Semantic Web community. Recent advances in translation-based
graph embedding methods for populating instance-level knowledge
graphs lead to promising new approaching for the ontology
population problem. However, unlike instance-level graphs, the majority
of relation facts in ontology graphs come with comprehensive
semantic \mbox{relations}, which often include the properties of transitivity and symmetry,
as well as hierarchical relations. These comprehensive relations are
often too complex for existing graph embedding methods, and direct
application of such methods is not feasible. Hence, we propose \ontovec,
a novel translation-based graph embedding method for ontology population.
\ontovec\ integrates two model components that effectively
characterize comprehensive relation facts in ontology graphs. The first
is the Component-specific Model that encodes concepts and relations into
low-dimensional embedding spaces without a loss of relational
properties; the second is the Hierarchy Model that performs focused learning
of hierarchical relation facts. Experiments on several well-known ontology
graphs demonstrate the promising capabilities of \ontovec\ in predicting
and verifying new relation facts. These promising results also make
possible significant improvements in related methods.
\end{abstract}

%% file: introduction.tex
\section{Introduction}
Ontology graphs are a special category of knowledge graphs that support and augment the Semantic Web with comprehensive and transportable machine understanding 
\cite{maedche2001learning}.
They store formal descriptions and specification of human knowledge in forms of relation facts (triples), making it semantically understandable and inferrable for the machine.
Unlike other instance-level knowledge graphs~\cite{zaniolo2017user} that define simple and casual labeled relations for specified entities, ontology graphs 
define a fixed set of specialized semantic relations among generalized concepts. 
Such semantic relations of ontologies are typically very comprehensive in terms of relational properties and form hierarchies, which we are going to discuss shortly.\par

Populating large ontologies has been a critical challenge to the Semantic \mbox{Web}.
In the past decade, several well-known ontology graphs have been created and widely utilized, including Yago~\cite{mahdisoltani2014yago3}, ConceptNet~\cite{speer2017conceptnet}, and DBpedia OWL~\cite{lehmann2015dbpedia}.
Although some of these graphs contain millions of relation facts, 
they still face the coverage and completeness issues that have been the subject of much research~\cite{quan2004automatic,mousavi2014mining}.
This is because enriching such large structures of expertise knowledge requires levels of 
intelligence and labor that is hardly affordable to humans.
Hence, some works have proposed to mine ontologies from text using parsing-based~\cite{culotta2004dependency,mousavi2014text,fundel2007relex} or fuzzy-logic-based~\cite{quan2004automatic,lau2009toward,widyantoro2001fuzzy} techniques.
However, in practice, these techniques are often limited by the lack of high-quality reference corpora that are required for the harvest of the dedicated domain knowledge.
Also, the precise recognition of relation facts for the ontology is another unsolved problem, since these relation facts are very high-level and are often not explicitly expressed in the corpora \cite{lau2009toward}.
Hence, these methods merely help populate some small ontology graphs in narrow domains such as gene ontologies and scholarly ontologies~\cite{cheng2004netaffx,quan2004automatic}, but they have not been successfully used to improve the completeness of these large cross-domain ontology graphs such as Yago and ConceptNet.\par
A more practical solution is to use translation-based graph embedding methods, which predict the missing relation facts using vector representations of the graph, without the need of additional information from any text corpus.
Specifically, given a triple $(s, r, t)$ such that $s$, $t$ denote the source and the target entities (or concepts), and $r$ denotes the edge that marks the relation between $s$ and $t$, then $s$ and $t$ are represented as two $k$-dimensional vectors $\mathbf{s}$ and $\mathbf{t}$, respectively.
An energy function $S_{r}(\mathbf{s},\mathbf{t})$ is used to measure the plausibility of the triple, which also implies the transformation $\mathbf{r}$ that characterizes $r$.
Therefore, new triples with high plausibility (or low energy) are often induced.
For example, TransE~\cite{bordes2013translating} uses the energy function $S_{r}(\mathbf{s},\mathbf{t})=\left \| \mathbf{s}+\mathbf{r}-\mathbf{t} \right \|$~\footnote{Hereafter, $\|\cdot\|$ means $l_1$ or $l_2$ norm unless specified.}, where $\mathbf{r}$ is characterized as a translation vector learnt from the latent connectivity patterns in the graph.
Other representative works, such as TransH~\cite{wang2014knowledge}, TransR~\cite{lin2015learning}, and TransD~\cite{ji2015knowledge} improve TransE by specializing
the encoding process for each relation type using a relation-specific projection on entities. \par

While these methods help enrich instance-level knowledge graphs,
they only focus on capturing the simple relations in instance-level knowledge graphs,
paying less attention to the comprehensive relations in ontology graphs.
In fact, relation facts in ontology graphs are often defined with relational properties, such as transitivity and symmetry, as well as form hierarchies.
A typical example is provided by {\em Is-A}, which is both transitive and hierarchical, and is the most frequently appearing semantic relation in ontologies.
We find that, in well-known ontology graphs, comprehensive relations usually comprise the majority: 85\% of the triples in Yago, 96\% of the triples in ConceptNet, and 47\% of the triples in DBpedia OWL enforce relational properties, while 60\%, 38\%, and 48\% of these triples are defined with hierarchical relations. 
However, existing \mbox{methods} fail to represent these comprehensive relations for several reasons:
(i) These methods at most use the same relation-specific projection in the energy function,
but fail to differentiate the components of triples.
Therefore, they are ill-posed to characterize triples with relational properties.
In fact, the encoding of a concept that serves as different components in such triples, i.e. either $s$ or $t$, must be differentiated so as to correctly preserve relational properties in the embedding spaces (as shown in Section~\ref{sect:pre}).
(ii) These methods also lack a learning phase that is dedicated to hierarchical relations. This also 
impairs the preciseness of embeddings.
We observe in our experiments that, above \mbox{limitations} largely hinder the effectiveness of existing methods for ontology graphs.\par

Therefore, to support ontology population more effectively, we propose \ontovec, a translation-based graph embedding model that specializes in characterizing the comprehensive semantic relations in ontology graphs.
\ontovec\ adopts two component models: the {\em Component-specific Model} which preserves the relational properties by applying component-specific projections on source and target concepts respectively, and the {\em Hierarchy Model} which performs an attentive learning process on hierarchical relations.
We evaluate our model with the tasks of relation prediction and relation verification, which respond respectively to the following two questions: (i) What relation should be added between two concepts? (ii) Is the predicted relation correct?
Experimental results on data sets extracted from Yago, ConceptNet, and DBpedia OWL show promising results and significant improvement on related methods. \par

The rest of the paper is organized as follows. We first discuss the related work, and then introduce our approach in the section that follows. After that we present the experimental evaluation, and conclude the paper in the last section.




%% file: related_work.tex
\inv
\section{Related Work}
\inv
In this section, we discuss three lines of works that are related to our topic.

\stitle{Ontology Population.}
Extensive human efforts are often put into the creation of ontology graphs.
Thus, ontology population aims at automatically extending the graphs with the missing relation facts.
A traditional strategy is to mine those facts from text corpora.
Many works rely on parsing-based techniques to harvest relation facts ~\cite{culotta2004dependency,mousavi2014text,wang2006automatic,giuliano2008instance,fundel2007relex}.
These approaches often construct hundreds or thousands of rules or parse-trees that are not reusable,
and human involvement is indispensable to filter the frequently generated conflict candidates.
Other works depend on fuzzy logic~\cite{quan2004automatic,lau2009toward,widyantoro2001fuzzy} to generate relation facts with uncertainty, which is more tractable than parsing-based techniques and do not generate conflict candidates.
However, identifying or summarizing the concepts from text still requires human intelligence.
Moreover, methods mentioned above suffer from the lack of reference corpora that are closely related to and highly cover the knowledge of the ontology.
Moreover, to associate the right contexts of the corpora with corresponding relation facts creates another major challenge, as semantic relations in ontologies are often specialized and are not explicitly expressed in the text.
Due to these issues, we have seen few successful applications of these traditional approaches in improving the coverage of large cross-domain ontology graphs like Yago and ConceptNet. 
These issues motivate us to consider the 
more flexible ``text-free'' methods based on translation-based graph embeddings.

\stitle{Translation-based Graph Embedding Methods.}
Recently, significant advancements have been made in learning translation-based embeddings for knowledge graphs.
To characterize a triple $(s, r, t)$, models of this family follow the common assumption that
$\mathbf{s}_r + \mathbf{r} \approx \mathbf{t}_r$,
where $\mathbf{s}_r$ and $\mathbf{t}_r$ are either the original vectors of $s$ and $t$, or the transformed vectors $f_{r}(s)$ and $f_{r}(t)$ under a 
certain transformation $f_{r}$ w.r.t. relation $r$.
The forerunner
TransE~\cite{bordes2013translating} sets $\mathbf{s}_r$ and $\mathbf{t}_r$ as the original $\mathbf{s}$ and $\mathbf{t}$.
Later works improve TransE by introducing
relation-specific projections on entities
to obtain different $\mathbf{s}_r$ and $\mathbf{t}_r$, including
projections on
relation-specific hyperplanes in TransH~\cite{wang2014knowledge},
linear transformations to multiple relation spaces in TransR~\cite{lin2015learning}, dynamic matrices in TransD~\cite{ji2015knowledge}, and other forms~\cite{jia2016locally,nguyenstranse}.
These variants of TransE specialize
the encoding process for different relations, therefore they often achieve better representations of triples than TransE.
Meanwhile translation-based models 
cooperate well with other models.
For example, variants of TransE are trained in joint 
with word embeddings to enable synthesized word embeddings with relational inferences~\cite{wang2014joint,zhong2015aligning},
and are combined with alignment models to help cross-lingual knowledge alignment~\cite{chen2017multi,chen2017akbc}. 
However, existing translation-based models are not able to preserve triples with relational properties in the embedding spaces, because they do not differentiate the encoding of concepts that serve as different components in these triples. They also fail to provide a proper learning process for hierarchical relations.
These are the major limitations we want to overcome.\par

On the other hand, to enrich the knowledge in graphs, translation-based models proceed with {\em entity prediction} that predicts missing entities for triples.
Since the candidate space of entities is extremely large, all these works seek to rank a set of candidates rather than acquiring the exact answers \cite{wang2014knowledge,lin2015learning,ji2015knowledge,jia2016locally,nguyenstranse}.
We instead proceed with relation prediction, which practically obtains the exact answers, as the relations in ontology graphs are not very diverse.

\stitle{Other Knowledge Graph Embedding Methods.} There are non-translation-based methods that learn graph embeddings.
UM~\cite{bordes2011learning}, SME~\cite{bordes2012joint} are simplified versions of TransE and TransR; LFM~\cite{jenatton2012latent} learns bilinear transformations among entities; {TADW}~\cite{yang2015network} learns context-based embeddings from random-walk generated contexts of the graphs (which is very similar to the recently introduced Rdf2Vec~\cite{ristoski2016rdf2vec}).
These methods do not explicitly embed relations, thus do not apply to our tasks.
Others include neural-based models SLM~\cite{collobert2008unified}, and NTN~\cite{socher2013reasoning} that were outperformed substantially by TransE and other translation-based methods on the tasks for populating instance-level knowledge graphs~\cite{bordes2013translating,wang2014knowledge,lin2015learning,ji2015knowledge}.
There are some which perform comparably with translation-based methods, but at the cost of much higher parameter complexity, such as RESCAL~\cite{nickel2011three}, and HolE~\cite{nickel2016holographic}. We choose to compare with these two popular methods as well.

%% file: model.tex
\section{Embedding Ontology Graphs}

In this section, we introduce the proposed method for learning ontology graph embeddings.
We begin with the formalization of ontology graphs.

\subsection{Preliminary}

An ontology is a graph $G(C, R)$ where $C$ is the set of concepts, and $R$ is the set of semantic relations.
$T = (s,r,t) \in G$ denotes a triple that represents a relation fact, 
for which $s, t \in C$ and $r \in R$.
Boldfaced $\mathbf{s}$, $\mathbf{r}$, $\mathbf{t}$ respectively represent the embedding vectors of source $s$, relation $r$, and target $t$.
Relations are further classified by $R=R_{tr} \cup R_{s} \cup R_h \cup R_{o}$, which respectively denote the sets of 
transitive, symmetric, hierarchical, and other simple relations.
We do not specify reflexive relations here because such relations can be easily model as a zero vector by any translation-based model.
$R_{tr}$ and $R_h$ thereof, are not required to be disjoint, while $R_o$ is disjoint with all the rest three.
For transitive relations, that is to say, given $r \in R_{tr}$, and three different concepts $c_1, c_2, c_3 \in C$, if $(c_1, r, c_2), (c_2, r, c_3) \in G$, then $(c_1, r, c_3) \in G$.
As for symmetric relations, that is to say, given $r \in R_s$, and two different concepts $c_1, c_2 \in C$, if $(c_1, r, c_2) \in G$, then $(c_2, r, c_1) \in G$.
As for hierarchical relations, we further divide them into  $R_h=R_r \cup R_c$ where $R_r$ denotes refinement relations that partition coarser concepts into finer ones,
and $R_c$ denotes coercion relations that 
group finer concepts to coarser ones~\cite{camossi2006multigranular,chen2016converting,chen2016sac}.

\subsection{Modeling}~\label{sect:pre}

\ontovec\ adopts two component models that learn on the two facets of the 
ontology graph:
the {\em Component-specific Model} (CSM) which encodes concepts and relations into low-dimensional embedding spaces without the loss of the relational properties,
and the {\em Hierarchy Model} (HM) which strengthens the learning process on hierarchical relations with an auxiliary energy.\par

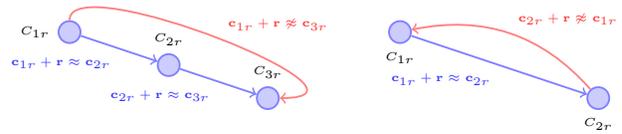
\begin{figure}
  \centering
  \resizebox {\columnwidth} {!} {
    \begin{tikzpicture}
      [nd/.style={circle,draw=blue!50,fill=blue!20,thick}]
      \node (c1r) at (-4, 0.5)  [nd,label=-180:{\tiny $C_{1r}$}] {};
      \node (c2r) at (-2.5, 0)  [nd,label=90:{\tiny $C_{2r}$}] {};
      \node (c3r) at (-1, -0.5) [nd,label=90:{\tiny $C_{3r}$}] {};

      \draw [->,draw=blue!50,thick] (c1r) to node [auto,swap,color=blue!80] {\tiny $\mathbf{c}_{1r}+\mathbf{r} \approx \mathbf{c}_{2r}$} (c2r);
      \draw [->,draw=blue!50,thick] (c2r) to node [auto,swap,color=blue!80] {\tiny $\mathbf{c}_{2r}+\mathbf{r} \approx \mathbf{c}_{3r}$} (c3r);
      \draw [->,draw=red!50,thick] (c1r) .. controls +(up:10mm) and +(right:20mm) .. node[auto,color=red!80] {\tiny $\mathbf{c}_{1r}+\mathbf{r} \not\approx \mathbf{c}_{3r}$} (c3r);

      \node (c4r) at (1, 0.5) [nd,label=-90:{\tiny $C_{1r}$}] {};
      \node (c5r) at (4, -0.5) [nd,label=-90:{\tiny $C_{2r}$}] {};
      \draw [->,draw=blue!50,thick] (c4r) to node [auto,swap,color=blue!80] {\tiny $\mathbf{c}_{1r}+\mathbf{r} \approx \mathbf{c}_{2r}$} (c5r);
      \draw [->,bend right,draw=red!50,thick] (c5r) to node[auto,swap,color=red!80] {\tiny $\mathbf{c}_{2r}+\mathbf{r} \not\approx \mathbf{c}_{1r}$} (c4r);
    \end{tikzpicture}
  }
  \caption{Depiction of the conflicts of the relation-specific projection for learning transitive relations (Case 1, left), and symmetric relations (Case 2, right).}\label{fig:conflict}
\end{figure}
\inv\inv

\subsubsection{Component-specific Model.}~\label{sect:CSM}

The reason that previous translation-based models fail to preserve relational properties is because the relation-specific projection $f_{r}$ place concepts involved in transitive or symmetric relations at conflict positions. Fig.~\ref{fig:conflict} depicts such conflicts, and a brief proof is given below:
\begin{itemize}[noitemsep]
\item \stitle{Case 1.} Consider $r \in R_{tr}$ and $c_1, c_2, c_3 \in C$ such that $(c_1, r, c_2)$, $(c_2, r, c_3)$, $(c_1, r, c_3) \in G$, where $c_1$, $c_2$, and $c_3$ are projected to $\mathbf{c}_{1r}$, $\mathbf{c}_{2r}$, and $\mathbf{c}_{3r}$ respectively by $f_{r}$. Then if $\mathbf{c}_{1r}+\mathbf{r}\approx \mathbf{c}_{2r}$ and $\mathbf{c}_{2r}+\mathbf{r}\approx \mathbf{c}_{3r}$ hold for the first and second triples, it is impossible for $\mathbf{c}_{1r}+\mathbf{r}\approx \mathbf{c}_{3r}$ to hold for the third triple, since $\mathbf{r} \neq 0$ (otherwise $\mathbf{r}$ does not provide a valid vector translation).
\item \stitle{Case 2.} Consider $r \in R_{s}$ and $c_1, c_2 \in C$ such that $(c_1, r, c_2), (c_2, r, c_1) \in G$, where $c_1$ and $c_2$ are projected to $\mathbf{c}_{1r}$ and $\mathbf{c}_{2r}$ respectively by $f_{r}$. Then it is not possible for both $\mathbf{c}_{1r}+\mathbf{r}\approx \mathbf{c}_{2r}$ and $\mathbf{c}_{2r}+\mathbf{r}\approx \mathbf{c}_{1r}$ to hold, since $\mathbf{r} \neq 0$.
\end{itemize}
Hence, to solve the conflicts in the above two cases, CSM provides two component-specific (and also relation-specific) projections to differentiate the encoding of the same concept that serves as different components in triples.
The general form of the energy function is given as below,
\small
\begin{equation*}~\label{eq:csm1}
S_d(T)=\left \| f_{1,r}(\mathbf{s})+\mathbf{r}-f_{2,r}(\mathbf{t}) \right \|
\end{equation*}
\normalsize
where $f_{1,r}$ and $f_{2,r}$ are respectively the component-specific projections for the source and the target concepts.
It is easy to show that the component-specific projections are able to solve the conflicts in learning the relational properties,
as $c_2$ in Case 1 is projected differently when it serves as the source of $(c_1, r, c_2)$ or the target of $(c_2, r, c_3)$, while both $c_1$ and $c_2$ in Case 2 can be learnt to be embedded in opposite positions respectively for $(c_1, r, c_2)$ and $(c_2, r, c_1)$ by the two projections.
Corresponding conclusion can be easily extended to cases with more than three relation facts via mathematical induction.\par

Besides measuring the plausibility (or the opposite: dissimilarity) of a given triple, $S_d$ is also the basis for predicting missing relation facts for an ontology. Given two concepts $s$ and $t$, we find the $r$ which leads to the lowest $S_d$. The forms of $f_{1,r}$ and $f_{2,r}$ are decided particularly by the techniques to differentiate the concept encoding under different contexts of relations.
In this paper, we adopt the relation-specific linear transformations \cite{lin2015learning}.
Hence, we have $f_{1,r}(\mathbf{s})=\mathbf{M}_{1,r}\mathbf{s}$ and $f_{2,r}(\mathbf{t})=\mathbf{M}_{2,r}\mathbf{t}$, such that $\mathbf{M}_{1,r},\mathbf{M}_{2,r} \in \mathbb{R}^{k \times k}$.
Other techniques like hyperplane projections, dynamic matrices, and bilinear transformations may also be considered, which we leave as future work.\par

The objective of CSM is to minimize the total $S_d$ energy of all triples.
To achieve more efficient learning, we import negative sampling to the learning process, which is widely applied in previous works~\cite{bordes2013translating,wang2014knowledge,lin2015learning,ji2015knowledge}.
Unlike these works that select negative samples on entities (or concepts), we perform negative sampling on semantic relations to better suit our tasks.
Then the complete energy function of CSM is defined as the following hinge loss,

\begin{align*}
\begin{split}
\small
S_{\mathrm{CSM}}(G)=\sum_{(s,r,t) \in G}[\left \| f_{1,r}(\mathbf{s})+\mathbf{r}-f_{2,r}(\mathbf{t}) \right \|\\
-\left \| f_{1,r}(\mathbf{s})+\mathbf{r'}-f_{2,r}(\mathbf{t}) \right \|+\gamma_1]_{+}
\normalsize
\end{split}
\end{align*}


for which $r'$ is a randomly sampled relation that does not hold between $s$ and $t$, $\gamma_1$ is a positive margin, and $[x]_{+}$ denotes the positive part 
of $x$ (i.e., $\max(x, 0)$).

\subsubsection{Hierarchy Model.}
For a hierarchical relation, we often have multiple finer concepts that apply this relation to a coarser one.
In this case, we appreciate a good representation where all the embeddings of the finer concepts converge closely in a tight neighborhood, which corresponds to 
low dissimilarity of the embedded relation. 
However, it is very likely for the learning process to spread out the embeddings of the finer concepts.
Because each of the finer concepts can participate in multiple relation facts, encoding of a concept in one relation fact can be easily interfered by that of many other relation facts.
This no doubt indicates low plausibility measures of the triples, and imprecise vector translation for the corresponding relations.
Therefore, HM is dedicated to converge closely the projected embeddings of every finer concepts for a hierarchical relation. \par
To facilitate the definition of the energy function, we first define a {\em refine} operator denoted as $\sigma$:
\begin{itemize}[noitemsep]
\item Given $r \in R_r$, $c\in C$,
then $\sigma(c, r)=\{c' | (c, r, c')\in G\}$
fetches all the finer concepts $c'$ that directly apply the refinement relation $r$ to the coarser $c$.
\item Given $r \in R_c$, $c\in C$,
then $\sigma(c, r)=\{c' | (c', r, c)\in G\}$
fetches all the finer concepts $c'$ that directly apply the coercion relation $r$ to the coarser $c$.
\end{itemize}
The energy function of HM is defined below,
\begin{align*}
\small
\begin{split}
S_{hm}(G)=\sum_{r\in R_r} \sum_{s\in C} \sum_{t\in \sigma(s,r)}
  \omega\left( f_{1,r} (\mathbf{s}) + \mathbf{r}, f_{2,r} (\mathbf{t}) \right)\\
  +\sum_{r\in R_c} \sum_{t\in C} \sum_{s\in \sigma(t,r)}
  \omega\left( f_{2,r} (\mathbf{t})-\mathbf{r}, f_{1,r} ( \mathbf{s}) \right)
\end{split}
\normalsize
\end{align*}
where $\omega$ is a function that monotonically increases w.r.t. the angle or the distance of the two argument vectors. In practice, $\omega$ can be easily implemented as cosine distance.\par
Negative sampling is imported to rewrite $S_{hm}$ as below,
\begin{align*}
\small
\begin{split}~\label{eq:hm}
S_{\mathrm{HM}}(G)=\sum_{r\in R_r}\sum_{s\in C}\sum_{t\in \sigma(s,r) \land t'\notin \sigma(s,r)} S_{hr}\\
+\sum_{r\in R_c} \sum_{t\in C}  \sum_{s\in \sigma(t,r) \land s'\notin \sigma(t,r)} S_{hc}
\end{split}
\normalsize
\end{align*}
such that $s'$ and $t'$ are negative samples of concepts, $S_{hr}$ and $S_{hc}$ are respectively the hinge loss for refinement and coercion relations defined as below, where $\gamma_2$ is a positive margin.
\small
\begin{equation*}~\label{eq:hr}
S_{hr}=[\omega\left( f_{1,r} (\mathbf{s}) + \mathbf{r}, f_{2,r} (\mathbf{t}) \right)-\omega\left( f_{1,r} (\mathbf{s}) + \mathbf{r}, f_{2,r} (\mathbf{t}') \right)+\gamma_2]_{+}
\end{equation*}
\normalsize
\small
\begin{equation*}~\label{eq:hc}
S_{hc}=[\omega\left( f_{2,r} (\mathbf{t})-\mathbf{r}, f_{1,r} ( \mathbf{s}) \right)-\omega\left( f_{2,r} (\mathbf{t}) - \mathbf{r}, f_{1,r} (\mathbf{s}') \right)+\gamma_2]_{+}
\end{equation*}
\normalsize
\input{alg1}

\input{tbl1}

Table~\ref{tbl:model_complexity} gives the model complexity of \ontovec\ and some related models in terms of parameter sizes.
We also give out the computational complexity of the relation prediction for a pair of concepts, which is the most frequent operation in our tasks.
Although \ontovec\ unavoidably increases the parameter sizes due to additional projections, it keeps the computational complexity of relation prediction at the same magnitude as TransR, which is lower than TransD.

\subsection{Learning Process}

The objective of learning \ontovec\ is to minimize the combined energy of $S_{\mathrm{CSM}}$ and $S_{\mathrm{HM}}$.
Meanwhile, norm constraints are enforced 
on embeddings and projections to prevent training from a trivial solution where vectors collapse to infinitely large~\cite{bordes2014semantic,chen2017multi,wang2014knowledge}.
Such constraints are conjuncted below.
\begin{align*}
\footnotesize
\begin{split}
\forall c \in C,\forall r \in R : \left \| \mathbf{c} \right \|\leq 1 \land \left \| f_{1,r}(\mathbf{c}) \right \|\leq 1 \land \left \| f_{2,r}(\mathbf{c}) \right \|\leq 1 \land \left \| \mathbf{r} \right \|\leq 2
\end{split}
\normalsize
\end{align*}
In the learning process, these constraints are quantified as soft constraints: 
\begin{align*}
\small
\begin{split}
S_{\mathrm{N}}(C,R)=\sum_{c \in C}([\left \| \mathbf{c} \right \|-1]_{+}+[\left \| f_{1,r}(\mathbf{c}) \right \|-1]_{+}\\
+[\left \| f_{2,r}(\mathbf{c}) \right \|-1]_{+})+\sum_{r \in R}[\left \| \mathbf{r} \right \|-2]_{+}
\end{split}
\normalsize
\end{align*}
Finally, learning \ontovec\ is realized by using batch stochastic gradient descent \mbox{(SGD)} \cite{needell2014stochastic} to minimize the joint energy function given as below,
\small
\begin{equation*}~\label{eq:joint}
J(\theta)=S_{\mathrm{CSM}}+\alpha_1 S_{\mathrm{HM}} + \alpha_2 S_{\mathrm{N}}
\end{equation*}
\normalsize
where $\alpha_1$ and $\alpha_2$ are two non-negative hyperparameters, and $\theta$ is the set of model parameters that include embedding vectors and projection matrices.
Empirically (as shown in~\cite{wang2014knowledge,lin2015learning}), $\alpha_2$ is assigned with a small value within (0, 1].
$\alpha_1$ is adjusted in experiments to weigh between the two component models.
Instead of directly updating $J$, the learning process optimizes $S_{\mathrm{CSM}}$ and $\alpha_1 S_{\mathrm{HM}}$ in separated groups of batches, and the batches from both groups are used to optimize $\alpha_2 S_{\mathrm{N}}$.
We initialize vectors by drawing from a uniform distribution on the unit spherical surface,
and initialize matrices using random orthogonal initialization \cite{saxe2013exact}.
The detailed optimization procedure is given in Algorithm~\ref{alg:training}.\par

%% file: alg1.tex
{\LinesNotNumbered

\begin{algorithm}[t]
  \caption{Learning procedure of On2Vec.}~\label{alg:training}
  \scriptsize
  \KwIn{Training set $G=\{(s,r,t\}$, hyperparameters $\alpha_1$ and $\alpha_2$, learning rate $\lambda$, batch size $b$}
   \KwOut{parameters $\theta$ for embedding vectors and projections}
   Randomly initialize $\theta$\;

   \While{training is not terminated}{
      $G_{\mathrm{CSM}} \leftarrow \mathsf{Sample}(G, b)$ \tcc*[r]{Sample size $b$.}
      $G_{\mathrm{HM}} \leftarrow B_{\mathrm{CSM}} \leftarrow B_{\mathrm{HM}} \leftarrow \emptyset$\;
      \While {$\left | G_{\mathrm{HM}} \right | < b$} {
         $c \leftarrow \mathsf{Sample}(c) \in C$\;
         $r \leftarrow \mathsf{Sample}(r) \in R_h$\;
         $G_{\mathrm{HM}} \leftarrow G_{\mathrm{HM}} \cup \sigma(c, r)$ \tcc*[r]{Truncate if $\left | G_{\mathrm{HM}} \right | \geq b$.}
      }
      \For {$T(s,r,t) \in G_{\mathrm{CSM}}$}{
         $T'(s,r',t) \leftarrow \mathsf{NegativeSample}(T)$\;
         $B_{\mathrm{CSM}} \leftarrow B_{\mathrm{CSM}} \cup \{(T, T')\}$ \tcc*[r]{Batch for CSM.}
      }
      \For {$T(s,r,t) \in G_{\mathrm{HM}}$}{
         \If {$r \in R_r$} {
         	\tcc{Negative sampling for a refinement relation.}
            $T'(s,r,t') \leftarrow \mathsf{NegativeSample}(T)$ \;
         }
         \Else {
         	\tcc{Negative sampling for a coercion relation.}
            $T'(s',r,t) \leftarrow \mathsf{NegativeSample}(T)$ \;
         }
         $B_{\mathrm{HM}} \leftarrow B_{\mathrm{HM}} \cup \{(T, T')\}$ \tcc*[r]{Batch for HM.}
      }
      $\theta \leftarrow \theta - \lambda \nabla S_{\mathrm{CSM}}(B_{\mathrm{CSM}})$\;
      $\theta \leftarrow \theta - \lambda \nabla \alpha_1 S_{\mathrm{HM}}(B_{\mathrm{HM}})$\;
      $B_c \leftarrow B_r \leftarrow \emptyset$ \tcc*[r]{Batch for soft-constraint.}
      \For {$(T,T') \in B_{\mathrm{CSM}} \cup B_{\mathrm{HM}}$} {
         $B_c \leftarrow B_c \cup \{s, s', t, t'\}$ \tcc*[r]{Concepts in triple batches.}
         $B_r \leftarrow B_r \cup \{r, r'\}$ \tcc*[r]{Relations in triple batches.}
      }
      $\theta \leftarrow \theta - \lambda \nabla \alpha_2 S_{N}(B_{c}, B_{r})$\;
}
\end{algorithm}
}
\inv

%% file: tbl1.tex
\begin{table*}[t!]
\centering
\begin{minipage}[t]{0.54\linewidth}
\centering
\caption{Model complexity: number of parameters for optimization, and the computational complexity for predicting a relation. $n_c$ and $n_r$ are numbers of concepts and relations, and $k$ is the dimensionality of embeddings.}
\label{tbl:model_complexity}	
\scriptsize			
\begin{tabular}{|c|c|c|}
\bhline
Model&\#Parameters&Complex. rel. predict.\\
\bhline
TransE&$\bigO{n_ck+n_rk}$&$\bigO{k+n_rk^2}$\\
TransH&$\bigO{n_ck+2n_rk}$&$\bigO{(3n_c+1)k+n_rk^2}$\\
TransR&$\bigO{n_ck+n_rk^2}$&$\bigO{n_ck^2+k+n_rk^2}$\\
TransD&$\bigO{n_ck+2n_rk}$&$\bigO{3n_ck^2+k+n_rk^2}$\\
\hline
On2Vec&$\bigO{n_ck+2n_rk^2}$&$\bigO{(n_c+1)k^2+k+n_rk^2}$\\
\bhline
\end{tabular}
\end{minipage}
\hfill
\begin{minipage}[t]{0.45\linewidth}
\centering
\caption{Statistics of the data sets. pct. prop. and pct. hier. are the percentages of triples defined with relational properties and hierachies.}
\label{tbl:stat}
\vspace{-1em}
{\scriptsize
\begin{tabular}{|c|c|c|c|c|}
\bhline
Data Set&DB3.6k&CN30k&YG15k&YG60k\\
\bhline
\#trip.&6,485&286,763&219,472&522,282\\
pct. prop.&47.39\%&96.89\%&45.69\%&85.58\%\\
pct. hier.&47.11\%&59.96\%&76.80\%&59.96\%\\
\#rel.&8&41&17&17\\
\#con.&3,625&29,564&14,887&56,910\\
\hline
\#train.&5,485&256,762&204,064&472,280\\
\#valid.&500&10,001&5,000&10,000\\
\#test.&500&20,000&10,400&40,000\\
\bhline
\end{tabular}
}
\end{minipage}
\hfill
\end{table*}

%% file: experiment.tex
\section{Experiments}

In this section, we evaluate \ontovec\  on two tasks that answer two important questions for ontology population: (i) Relation prediction: what is the relation to be added between a given pair of concepts? (ii) Relation verification: is a candidate relation fact correct or not?\par
The baselines that we compare against include the representative translation-based embedding methods TransE, TransH, TransR, and TransD \cite{bordes2013translating,wang2014knowledge,lin2015learning,ji2015knowledge}, and neural methos RESCAL and HolE \cite{nickel2011three,nickel2016holographic}.
Experimental results are reported on four data sets extracted from DBpedia, ConceptNet, and Yago, for which comprehensive relation types have been predefined.
Statistics of the data sets are shown in Table~\ref{tbl:stat}.
All the meta relations that assign URIs and system timestamps are removed during the preparation of the data sets.
To simplify the experiments, transitive relations are limited to four-hops. Relation facts for extra hops are hence discarded.
Since DBpedia provides both ontology and instance-level graphs, we keep only the ontology view to obtain DB3.6k.
CN30k and YG15k are extracted from English versions of ConceptNet and Yago respectively.
These two graphs match the number of nodes with WN18 and FB15k respectively, which are two commonly-used instance-level graphs in related works~\cite{bordes2013translating,wang2014knowledge,lin2015learning,ji2015knowledge,bordes2011learning,bordes2012joint,yang2015network}.
YG60k is a much larger data set that is about half of the entire English-version Yago after data cleaning. 
Each data set is randomly partitioned into training, validation, and test sets.
\input{tbl2}
\input{fig1}

\subsection{Relation Prediction}~\label{sect:rel_pred}

This task aims at extending an ontology graph by predicting the missing relations for given concept pairs. \par

\stitle{Evaluation Protocol.} We evaluate our approach by way of held-out evaluation~\cite{wang2014joint,lin2016neural}.
Each model is trained on the training set that represents the known ontology.
Then, for each case in the test set, given the source and target concepts, the model predicts 
the relation that leads to the lowest dissimilarity score $S_d$ defined in Section~\ref{sect:CSM}.
To evaluate with controlled variables, on each data set, we employ the same configuration for every models.
On DB3.6k, we fix dimensionality $k=25$, margin $\gamma_1=2.0$, learning rate $\lambda=0.005$, $\alpha_2=0.5$, and $l_1$ norm.
CN30k and YG15k shares the configuration as $k=50$, $\gamma_1=0.5$, $\lambda=0.001$, $\alpha_2=0.5$, and $l_2$ norm.
Lastly, we use $k=100$, $\gamma_1=0.5$, $\lambda_1=0.001$, $\alpha_2=0.5$, and $l_2$ norm.
$\gamma_2=0.5$ is configured for \ontovec.
To test the effect of HM, we also provide two versions of \ontovec. One version (\ontovec\ w/ HM) is set with $\alpha_1=0.75$, which is empirically decided via the hyperparameter study in Section~\ref{sect:case}. The other version (\ontovec\ w/o HM) nullifies HM by setting $\alpha_1=0$.
To enable batch sampling for HM, we implement the $\sigma$ function for hierarchical relation facts using hash trees.
The learning process is stopped once the accuracy on the validation set stops improving. \par

\stitle{Results.} The overall accuracy is reported per data set in Table~\ref{tbl:predict}.
On each data set, we also aggregate respectively the accuracy on the test cases with relational properties, as well as the accuracy on those with hierarchical relations.
We discover that, TransE, though has performed well on encoding instance-level knowledge graphs~\cite{bordes2013translating}, receives unsatisfactory results on predicting the comprehensive ontology relations.
By 
learning each relation type on a different hyperplane, TransH notably solves the problem of TransE, but appears to fail on CN30k where the candidate space is larger than other graphs.
TransR and TransD provide more robust characterization of relations than TransH, especially in TransR where relation-specific projections are implemented as linear transformations.
However, the overall performance of both TransR and TransD is impaired by the two types of comprehensive relations. 
For neural models, HolE adapts better on the smaller DB3.6k data set, while it is at most comparable to TransR and TransD on larger ones,
and RESCAL is less successful on all settings.
As expected, 
\ontovec\ greatly outperforms the above baseline methods, regardless of whether HM is enabled or not.
The \ontovec\ with HM thereof, outperforms the best runner-up baselines respectively in all settings by 3.72\%${\sim}$10.52\% of overall accuracy, 4.09\%${\sim}$11.69\% of accuracy on cases with relational properties, and 7.97\%${\sim}$14.24\% of accuracy on cases with hierarchical relations.
We also discover that, when HM is enabled, it leverages the accuracy on hierarchical relations by up to 12.75\%, and overall accuracy by up to 7.27\%, and 
does not noticeably cause interference to the prediction for cases with relational properties.
Though, the advantage of CSM alone (i.e. \ontovec\ w/o HM) is still significant over the baselines.
Since the relation prediction accuracy of \ontovec\ is close to 90\% on all four data sets, this indicates that \ontovec\ achieves a promising level of performance in populating ontology graphs,
and it is effective on both small and large graphs.\par

We also perform precision-recall analysis on the two Yago data sets on translation-based models.
To do so, we calculate the dissimilarity scores $S_d$ (Equation~\ref{eq:csm1}) for 
the possible predictions of each test case, and select those that are not ranked behind the correct prediction.
Then a threshold is initiated as the minimum dissimilarity score.
The answer set is inserted with predictions for which the dissimilarity scores fall below the threshold, and the answer set grows along with the increasing of the threshold, until all correct predictions are inserted.
Therefore, we obtain the precision-recall curves in Fig.~\ref{fig:prc},
for which the area under curve is reported as: (i) For YG15k, \ontovec\ w/ HM: \textbf{0.9138}; \ontovec\ w/o HM: 0.8938; TransE: 0.0457; TransH: 0.4973; TransD: 0.8386; TransR: 0.8587. (ii) For YG60k, \ontovec\ w/ HM: \textbf{0.9005}; \ontovec\ w/o HM: 0.8703; TransE: 0.0313; TransH: 0.6688; TransD: 0.7275; TransR: 0.8372.
This further indicates that \ontovec\ achieves better performance than other baselines, and HM improves the performance of On2Vec with CSM alone.

\subsection{Relation Verification}

\input{tbl3}

Relation verification aims at judging whether a relation marked between two concepts is correct or not.
It produces a classifier that helps to verify the candidate relation facts.\par
\stitle{Evaluation Protocol.} Because this is a binary classification problem that needs positive and negative cases, we use a complete data set as the positive cases.
Then, following the approach of \cite{socher2013reasoning}, we corrupt the data set to create negative cases.
In detail, a negative case is created by (i) randomly replacing the relation of a positive case with another relation, or (ii) randomly assign a relation to a pair of unrelated concepts.
Options (i) and (ii) respectively contribute negative cases that are as many as 100\% and 50\% of positive cases.
We perform a 10-fold cross-validation.
Within each fold, embeddings and the classifier are trained on the training data, and the classifier is evaluated on the remaining validation data.\par
We use a threshold-based 
classifier, which is similar to the one for triple alignment verification in \cite{chen2017multi}.
This simple classifier adequately relies on how precisely each model preserves the structure of the ontology graph in the embedding space.
In detail, for each case, we calculate its dissimilarity score $S_d$ (Section~\ref{sect:CSM}). 
The classifier then finds a threshold $\tau$ such that $S_d < \tau$ implies positive, otherwise negative.
The value of $\tau$ is determined to maximize the accuracy on the training data of each fold.\par
We carry forward the corresponding configurations from the last experiment, in order to show the performance of each model under controlled variables.\par
\input{tbl4}
\stitle{Results.} We aggregate the mean accuracy for the two categories of comprehensive relation facts as well as the overall accuracy for each setting. The results are shown in Table~\ref{tbl:classification}, which has a maximum standard deviation of 0.005 in cross-validation for each setting.
Thus, the results are statistically sufficient to reflect the performance of classifiers.
Both versions of \ontovec\ again outperform the other models, especially on comprehensive relation facts.
On all four data sets, \ontovec\ outperforms the best runner-up baselines by 2.98\%${\sim}$9.67\% of overall accuracy, 2.02\%${\sim}$12.57\% of accuracy for cases with relational properties, and 1.29${\sim}$8.15\% of accuracy on hierarchical relations.
This indicates that \ontovec\ precisely encodes the ontology graph structures, and provides much accurate plausibility measurement to decide the correctness of unknown triples.
We also discover that, \ontovec\ trained with HM has a drop of accuracy for up to 0.8\% on cases with relational properties from CN30k and YG15k.
This is likely due to that the auxiliary learning process for hierarchical relations causes minor interference to the characterization of relational properties, while HM leverages the accuracy on hierarchical relations of these two data sets by at least 1.81\%, and the overall accuracy by 0.82\%${\sim}$3.83\%.
This indicates that HM is helpful in relation verification.

\subsection{Case Study}~\label{sect:case}
Lastly, we provide some case studies 
on hyperparameter values, and some examples of relation prediction.
\subsubsection{Hyperparameter study}
We examine the hyperparameter $\alpha_1$, which is the trade-off between CSM and HM.
The result based on relation prediction on YG15k is shown in Fig.~\ref{fig:alpha}.
As we can see, although enabling HM with even a small value of $\alpha_1$ can noticeably leverage the performance of \ontovec, the influence of different values of $\alpha_1$ is not very notable,
and the accuracy does not always go up along with the higher $\alpha_1$.
In practice, $\alpha_1$ may be fine-tuned for marginal improvement, while $\alpha_1=0.75$ can be empirically selected.

\subsubsection{Examples of relation prediction}
Relation prediction is also performed for the complete data set of CN30k and YG60k.
To do so, we randomly select 20 million pairs of unlinked concepts from these two data sets, and rank all the predictions based on the dissimilarity score $S_d$. Then top-ranked predictions are selected.
Human evaluation is used in this procedure, since there is no ground truth for the relation facts that are not pre-existing.
Like previous works~\cite{lin2016neural,zeng2015distant}, we aggregate $P@200$, i.e. the precision on the 200 predictions with highest confidence, which results in 73\% and 71\% respectively.
Some examples of top-ranked predictions are shown in Table~\ref{tbl:new_rel}.

%% file: tbl2.tex
\begin{table*}[t!]
\centering
\caption{Accuracy of Relation Prediction (\%). prop. means with properties, hier. means hierarchical relations.}~\label{tbl:predict}
\label{tbl:rel_predict}
\vspace{-1em}
\scriptsize
\begin{tabular}{|c|ccc|ccc|ccc|ccc|}
\bhline
Data Sets&\multicolumn{3}{c|}{DB3.6k}&\multicolumn{3}{c|}{CN30k}&\multicolumn{3}{c|}{YG15k}&\multicolumn{3}{c|}{YG60k}\\
\bhline
Rel Type&prop.&hier.&overall&prop.&hier.&overall&prop.&hier.&overall&prop.&hier.&overall\\
\bhline
TransE&8.40&8.71&13.31&5.09&3.21&8.01&2.03&0.56&0.20&0.02&0.00&0.16\\
TransH&47.55&47.83&50.80&13.8&7.29&13.66&65.53&61.57&66.27&62.92&43.79&59.78\\
TransD&50.40&57.98&80.74&72.34&76.18&77.67&74.42&75.60&77.77&72.39&66.18&73.23\\
TransR&68.14&71.72&78.32&79.32&84.37&80.56&79.74&79.56&79.81&77.40&71.19&78.22\\
RESCAL&29.70&35.65&36.19&55.39&56.06&54.46&58.88&54.50&59.07&52.36&53.16&58.51\\
HolE&82.76&81.68&89.63&79.21&80.99&77.71&76.78&75.20&79.13&73.69&74.47&78.10\\
\hline
O2V w/ HM&86.46&\textbf{89.65}&\textbf{93.35}&\textbf{88.99}&\textbf{96.05}&\textbf{89.21}&\textbf{88.88}&\textbf{89.36}&\textbf{88.75}&\textbf{89.09}&\textbf{88.71}&\textbf{88.74}\\
O2V w/o HM&\textbf{86.85}&86.06&90.69&85.58&95.07&86.01&85.87&83.98&84.29&80.57&75.96&81.47\\
\bhline
\end{tabular}
\end{table*}

%% file: fig1.tex
\noindent
\inv
\begin{figure}[t!]
\centering
\begin{minipage}[]{0.48\textwidth}
\centering
\includegraphics[width=1.0\textwidth]{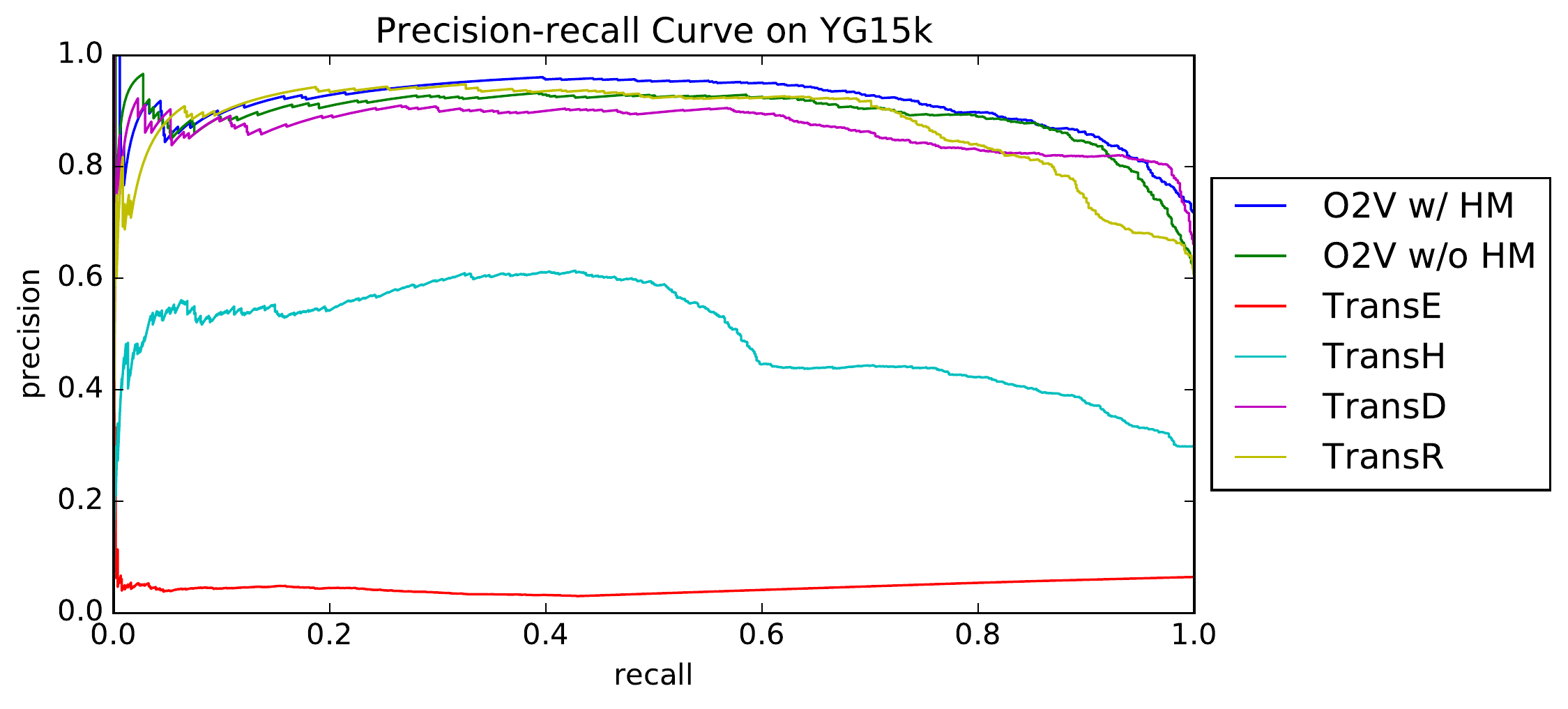}
\includegraphics[width=1.0\textwidth]{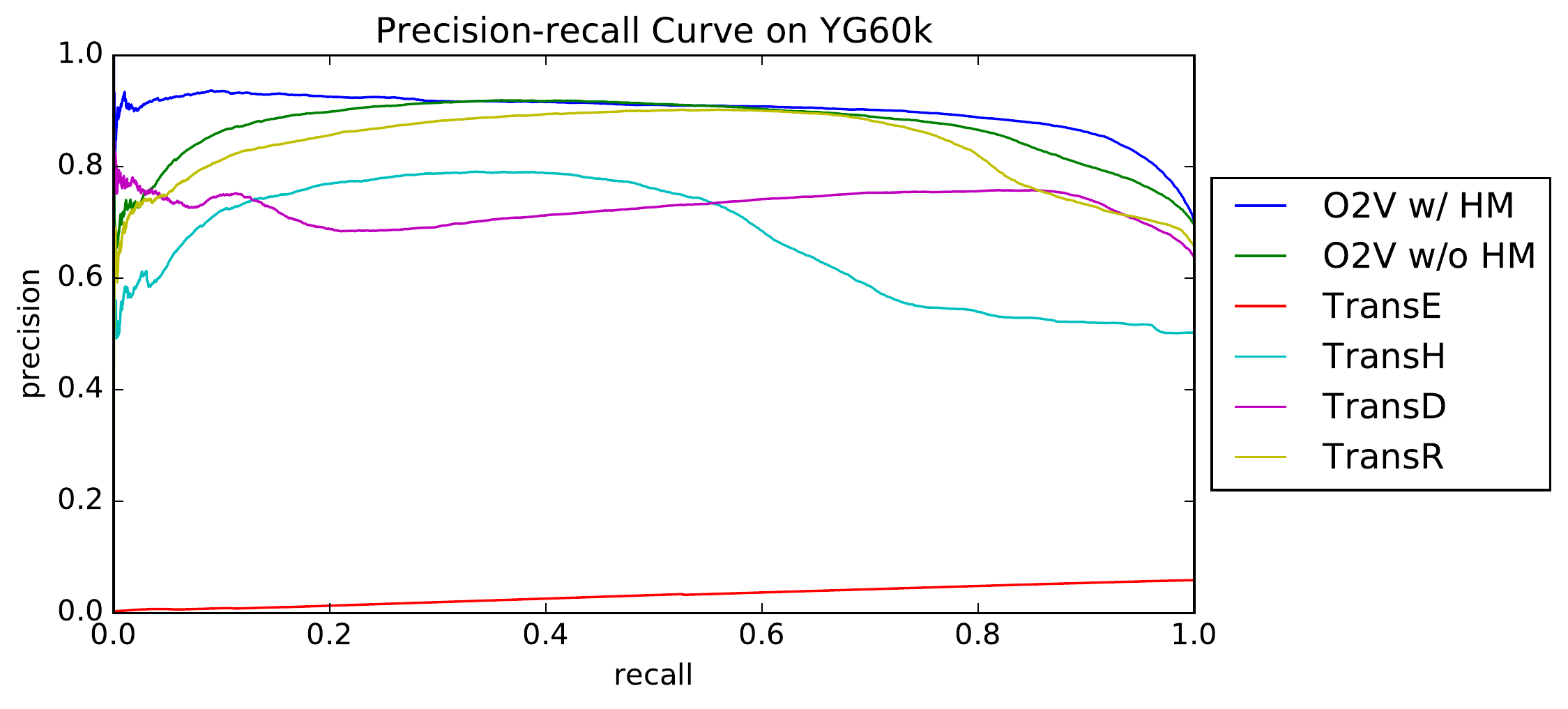}
\caption{Precision-recall curves for relation prediction on YG15k and YG60k. }~\label{fig:prc}
\end{minipage}
\vspace{-1.5em}
\end{figure}

%% file: tbl3.tex
\begin{table*}[t]
\centering
\caption{Accuracy of relation verification (\%). prop. means with properties, hier. means hierarchical relations.}~\label{tbl:classification}
\label{tbl:rel_predict}
\vspace{-1em}
\scriptsize
\begin{tabular}{|c|ccc|ccc|ccc|ccc|}
\bhline
Data Sets&\multicolumn{3}{c|}{DB3.6k}&\multicolumn{3}{c|}{CN30k}&\multicolumn{3}{c|}{YG15k}&\multicolumn{3}{c|}{YG60k}\\
\bhline
Rel Type&prop.&hier.&overall&prop.&hier.&overall&prop.&hier.&overall&prop.&hier.&overall\\
\bhline
TransE&67.49&71.44&67.57&69.14&18.23&51.85&58.73&62.69&69.09&60.92&61.30&66.89\\
TransH&72.88&82.06&69.71&93.40&86.16&94.17&69.24&72.96&89.20&66.47&71.62&88.81\\
TransD&76.79&81.11&74.44&91.63&84.20&93.36&65.63&70.58&88.01&61.76&71.08&86.34\\
TransR&77.11&86.82&73.76&85.83&52.01&74.73&71.80&72.73&88.63&71.92&71.09&87.77\\
RESCAL&75.30&74.61&76.20&70.41&75.64&72.28&68.76&67.30&72.29&69.36&69.16&76.21\\
HolE&82.89&79.23&85.90&90.31&91.43&91.18&78.31&77.10&86.88&71.22&70.80&87.67\\
\hline
O2V w/ HM&\textbf{95.46}&\textbf{94.97}&\textbf{95.57}&97.19&\textbf{95.54}&\textbf{98.04}&80.33&\textbf{78.39}&\textbf{93.29}&\textbf{74.92}&\textbf{73.36}&\textbf{91.79}\\
O2V w/o HM&91.94&91.15&91.74&\textbf{97,99}&93.73&96.51&\textbf{81.01}&74.30&91.12&73.72&72.93&90.97\\
\bhline
\end{tabular}
\end{table*}

%% file: tbl4.tex
\begin{table}[t]
\centering
\caption{Examples of top-ranked new relation facts. The italic ones are conceptually close. The rest are correct.}~\label{tbl:new_rel}
\scriptsize
\begin{tabular}{|l|}
\bhline
CN30k\\
\bhline
$<$Offer, Entails, Degree$>$\\
$<$Offer, Entails, Decide$>$\\
$<$State, IsA, Boundary$>$\\
$<$National\_Capital, IsA, Boundary$>$\\
$<${\em Get\_in\_line, HasFirstSubevent, Pay}$>$\\
$<$Convert, SimilarTo, Transform$>$\\
$<$Person, ReceivesAction, Hint$>$\\
$<${\em Stock, Entails, Receive}$>$\\
$<${\em Evasion, HasContext, Physic}$>$\\
\bhline
YG60k\\
\bhline
$<$Luisa\_de\_Guzm$\acute{a}$n, isMarriedTo, John\_IV\_of\_Portugal$>$\\
$<${\em Georgetown, isLocatedIn, South\_Carolina}$>$\\
$<$Gmina\_pomiech$\acute{o}$wek, isLocatedIn, Gmina\_Konstancin$>$\\
$<$$\ddot{O}$rebro\_Airport, isLocatedIn, Karlskoga$>$\\
$<${\em Horgen, isLocatedIn, B$\ddot{u}$lach\_District}$>$\\
$<$Luxor\_International\_Airport, isConnectedTo,\\
Daqing\_Sartu\_Airport$>$\\
$<$Akron, isLocatedIn, Ohio$>$\\
$<$Curtis\_guild\_Jr, hasGender, Male$>$\\
$<$Aalbach, isLocatedIn, Europe$>$\\
\bhline
\end{tabular}
\vspace{-1.5em}
\end{table}

\begin{figure}[t]
  \centering
  \includegraphics[width=0.48\textwidth]{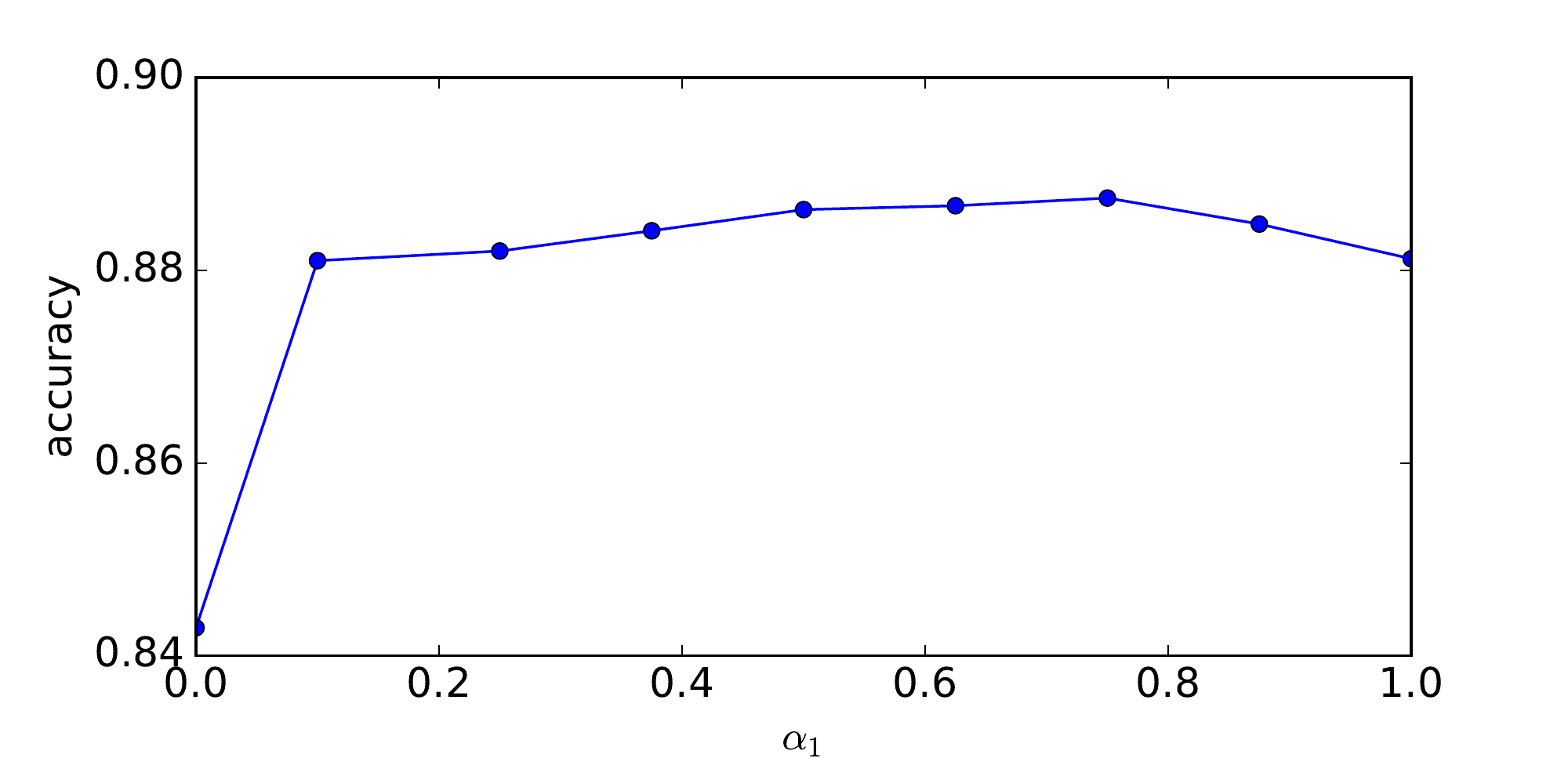}\\
  \vspace{-1.5em}
  \caption{Choices of $\alpha_1$ values and corresponding accuracy of relation prediction on YG15k.}\label{fig:alpha}
\end{figure}
\inv

%% file: conclusion.tex
\section{Conclusion}
\inv
This paper proposes a greatly improved translation-based graph embedding method that helps ontology population by way of relation prediction.
The proposed \ontovec\ model can effectively address the learning issues on the two categories of comprehensive semantic relations in ontology graphs, and improves previous methods using two dedicated component models.
Extensive experiments on four data sets show promising capability of \ontovec\ on predicting and verifying relation facts.\par

The results here are very encouraging, but we also point out opportunities for further work and improvements. In particular, we
should explore the effects of other possible forms of component-specific projections, such as dynamic mapping matrices and bilinear mappings.
Encoding other information such as the domain and range information of concepts may also improve the precision of our tasks.
More advanced applications may also be developed using \ontovec\, such as ontology-boosted question answering.
Jointly training \ontovec\ with alignment models \cite{chen2017learning}
is another meaningful direction since it provides 
a more generic embedding model that helps populating and aligning multilingual ontology graphs.

%% file: on2v_sdm.bbl
\begin{thebibliography}{10}

\bibitem{bordes2012joint}
{\sc A.~Bordes, X.~Glorot, et~al.}, {\em Joint learning of words and meaning
  representations for open-text semantic parsing}, in AISTATS, 2012.

\bibitem{bordes2014semantic}
{\sc A.~Bordes, X.~Glorot, et~al.}, {\em A semantic matching energy function
  for learning with multi-relational data}, Machine Learning, 94 (2014).

\bibitem{bordes2013translating}
{\sc A.~Bordes, N.~Usunier, et~al.}, {\em Translating embeddings for modeling
  multi-relational data}, in NIPS, 2013.

\bibitem{bordes2011learning}
{\sc A.~Bordes, J.~Weston, et~al.}, {\em Learning structured embeddings of
  knowledge bases}, in AAAI, 2011.

\bibitem{camossi2006multigranular}
{\sc E.~Camossi, M.~Bertolotto, et~al.}, {\em A multigranular object-oriented
  framework supporting spatio-temporal granularity conversions}, IJGIS, 20
  (2006), pp.~511--534.

\bibitem{chen2016converting}
{\sc M.~Chen, S.~Gao, et~al.}, {\em Converting spatiotemporal data among
  heterogeneous granularity systems}, in FUZZ-IEEE, 2016.

\bibitem{chen2016sac}
{\sc M.~Chen, S.~Gao, et~al.}, {\em Converting spatiotemporal data among
  multiple granularity systems}, in SAC, 2016.

\bibitem{chen2017multi}
{\sc M.~Chen, Y.~Tian, et~al.}, {\em Multilingual knowledge graph embeddings
  for cross-lingual knowledge alignment}, IJCAI,  (2017).

\bibitem{chen2017learning}
{\sc M.~Chen and C.~Zaniolo}, {\em Learning multi-faceted knowledge graph
  embeddings for natural language processing}, IJCAI,  (2017).

\bibitem{chen2017akbc}
{\sc M.~Chen, T.~Zhou, et~al.}, {\em Multi-graph affinity embeddings for
  multilingual knowledge graphs}, in AKBC, 2017.

\bibitem{cheng2004netaffx}
{\sc J.~Cheng, S.~Sun, et~al.}, {\em Netaffx gene ontology mining tool: a
  visual approach for microarray data analysis}, Bioinformatics, 20 (2004).

\bibitem{collobert2008unified}
{\sc R.~Collobert and J.~Weston}, {\em A unified architecture for natural
  language processing: Deep neural networks with multitask learning}, in ICML,
  2008.

\bibitem{culotta2004dependency}
{\sc A.~Culotta and J.~Sorensen}, {\em Dependency tree kernels for relation
  extraction}, in ACL, 2004.

\bibitem{fundel2007relex}
{\sc K.~Fundel, R.~K{\"u}ffner, et~al.}, {\em {RelEx}--{Relation} extraction
  using dependency parse trees}, Bioinformatics, 23 (2007).

\bibitem{giuliano2008instance}
{\sc C.~Giuliano and A.~Gliozzo}, {\em Instance-based ontology population
  exploiting named-entity substitution}, in COLING, 2008.

\bibitem{jenatton2012latent}
{\sc R.~Jenatton, N.~L. Roux, et~al.}, {\em A latent factor model for highly
  multi-relational data}, in NIPS, 2012.

\bibitem{ji2015knowledge}
{\sc G.~Ji, S.~He, et~al.}, {\em Knowledge graph embedding via dynamic mapping
  matrix}, in ACL, 2015.

\bibitem{jia2016locally}
{\sc Y.~Jia, Y.~Wang, et~al.}, {\em Locally adaptive translation for knowledge
  graph embedding}, in AAAI, 2016.

\bibitem{lau2009toward}
{\sc R.~Y. Lau, D.~Song, et~al.}, {\em Toward a fuzzy domain ontology
  extraction method for adaptive e-learning}, TKDE, 21 (2009).

\bibitem{lehmann2015dbpedia}
{\sc J.~Lehmann, R.~Isele, et~al.}, {\em Dbpedia--a large-scale, multilingual
  knowledge base extracted from {Wikipedia}}, Semantic Web, 6 (2015).

\bibitem{lin2015learning}
{\sc Y.~Lin, Z.~Liu, et~al.}, {\em Learning entity and relation embeddings for
  knowledge graph completion.}, in AAAI, 2015.

\bibitem{lin2016neural}
{\sc Y.~Lin, S.~Shen, et~al.}, {\em Neural relation extraction with selective
  attention over instances}, in ACL, 2016.

\bibitem{maedche2001learning}
{\sc A.~Maedche and S.~Staab}, {\em Learning ontologies for the semantic web},
  in ISWC, 2001.

\bibitem{mahdisoltani2014yago3}
{\sc F.~Mahdisoltani, J.~Biega, et~al.}, {\em Yago3: A knowledge base from
  multilingual {Wikipedias}}, in CIDR, 2015.

\bibitem{mousavi2014text}
{\sc H.~Mousavi, M.~Atzori, et~al.}, {\em Text-mining, structured queries, and
  knowledge management on web document corpora}, {SIGMOD} Record, 43 (2014).

\bibitem{mousavi2014mining}
{\sc H.~Mousavi, S.~Gao, , et~al.}, {\em Mining semantics structures from
  syntactic structures in web document corpora}, International Journal of
  Semantic Computing, 8 (2014).

\bibitem{needell2014stochastic}
{\sc D.~Needell, R.~Ward, et~al.}, {\em Stochastic gradient descent, weighted
  sampling, and the randomized kaczmarz algorithm}, in NIPS, 2014.

\bibitem{nguyenstranse}
{\sc D.~Q. Nguyen, K.~Sirts, et~al.}, {\em Stranse: a novel embedding model of
  entities and relationships in knowledge bases}, in NAACL HLT, 2016.

\bibitem{nickel2016holographic}
{\sc M.~Nickel, L.~Rosasco, et~al.}, {\em Holographic embeddings of knowledge
  graphs}, in AAAI, 2016.

\bibitem{nickel2011three}
{\sc M.~Nickel, V.~Tresp, et~al.}, {\em A three-way model for collective
  learning on multi-relational data}, in ICML, 2011.

\bibitem{quan2004automatic}
{\sc T.~T. Quan, S.~C. Hui, et~al.}, {\em Automatic generation of ontology for
  scholarly semantic web}, in ISWC, 2004.

\bibitem{ristoski2016rdf2vec}
{\sc P.~Ristoski and H.~Paulheim}, {\em Rdf2vec: Rdf graph embeddings for data
  mining}, in ISWC, 2016.

\bibitem{saxe2013exact}
{\sc A.~M. Saxe, J.~L. McClelland, et~al.}, {\em Exact solutions to the
  nonlinear dynamics of learning in deep linear neural networks}, ICLR,
  (2014).

\bibitem{socher2013reasoning}
{\sc R.~Socher, D.~Chen, et~al.}, {\em Reasoning with neural tensor networks
  for knowledge base completion}, in NIPS, 2013.

\bibitem{speer2017conceptnet}
{\sc R.~Speer, J.~Chin, et~al.}, {\em Conceptnet 5.5: An open multilingual
  graph of general knowledge.}, in AAAI, 2017.

\bibitem{wang2006automatic}
{\sc T.~Wang, Y.~Li, K.~Bontcheva, et~al.}, {\em Automatic extraction of
  hierarchical relations from text}, in ESWC, 2006.

\bibitem{wang2014joint}
{\sc Z.~Wang, J.~Zhang, et~al.}, {\em Knowledge graph and text jointly
  embedding.}, in EMNLP, 2014.

\bibitem{wang2014knowledge}
{\sc Z.~Wang, J.~Zhang, et~al.}, {\em Knowledge graph embedding by translating
  on hyperplanes}, in AAAI, 2014.

\bibitem{widyantoro2001fuzzy}
{\sc D.~H. Widyantoro and J.~Yen}, {\em A fuzzy ontology-based abstract search
  engine and its user studies}, in FUZZ-IEEE, 2001.

\bibitem{yang2015network}
{\sc C.~Yang, Z.~Liu, et~al.}, {\em Network representation learning with rich
  text information}, in IJCAI, 2015.

\bibitem{zaniolo2017user}
{\sc C.~Zaniolo, S.~Gao, et~al.}, {\em User-friendly temporal queries on
  historical knowledge bases}, Information and Computation,  (2017).

\bibitem{zeng2015distant}
{\sc D.~Zeng, K.~Liu, et~al.}, {\em Distant supervision for relation extraction
  via piecewise convolutional neural networks.}, in EMNLP, 2015.

\bibitem{zhong2015aligning}
{\sc H.~Zhong, J.~Zhang, et~al.}, {\em Aligning knowledge and text embeddings
  by entity descriptions}, in EMNLP, 2015.

\end{thebibliography}
